\documentclass[11pt]{article}

\usepackage{multirow}
\usepackage{amssymb,amsmath,amsthm}
\usepackage{color}
\usepackage{graphicx}
\usepackage{tikz}
\usepackage{epstopdf}
\usepackage{romannum}
\usepackage{algorithm}
\usepackage{subcaption}
\usepackage{graphicx}
\usepackage{float}
\usepackage{multicol}
\usepackage[noend]{algpseudocode}
\usepackage{romannum}

\def\oneb{{\bf 1}}

\def\B{{\cal B}}

\def\R{{\mathbb R}}

\def\al{\alpha}
\def\d{\delta}
\def\D{\Delta}
\def\e{\epsilon}
\def\g{\gamma}

\def\l{\lambda}
\def\L{\Lambda}

\def\OM{\Omega}
\def\r{\rho}
\def\s{\sigma}
\def\SI{\Sigma}
\def\t{\tau}
\def\th{\theta}

\def\cb{\bar{c}}

\def\kb{\bar{k}}

\def\ph{\hat{p}}

\def\nai{n \ap \infty}
\def\tai{t \ap \infty}

\def\ap{\rightarrow}

\def\seq{\subseteq}

\def\bz{{\bf 0}}

\def\fa{\; \forall}

\def\st{\mbox{ s.t. }}

\def\nm{\Vert}

\renewcommand{\iff}{\mbox{$\; \; \Longleftrightarrow \; \;$}}
\renewcommand{\and}{\mbox{$\wedge$}}

\newcommand{\bc}{\begin{center}}
\newcommand{\ec}{\end{center}}
\newcommand{\be}{\begin{equation}}
\newcommand{\ee}{\end{equation}}
\newcommand{\bd}{\begin{displaymath}}
\newcommand{\ed}{\end{displaymath}}
\newcommand{\ba}{\begin{array}}
\newcommand{\ea}{\end{array}}
\newcommand{\ben}{\begin{enumerate}}
\newcommand{\een}{\end{enumerate}}
\newcommand{\bit}{\begin{itemize}}
\newcommand{\eit}{\end{itemize}}
\newcommand{\beq}{\begin{eqnarray}}
\newcommand{\eeq}{\end{eqnarray}}
\newcommand{\btab}{\begin{tabular}}
\newcommand{\etab}{\end{tabular}}
\newcommand{\bfig}{\begin{figure}}
\newcommand{\efig}{\end{figure}}
\newcommand{\btp}{\begin{tikzpicture}}
\newcommand{\etp}{\end{tikzpicture}}

\newcommand{\argmin}{\operatornamewithlimits{arg min}}

\newcommand{\nmm}[1]{ \nm #1 \nm }
\newcommand{\nmeu}[1]{ \nm #1 \nm_2 }
\newcommand{\nmeusq}[1]{ \nm #1 \nm_2^2 }

\newcommand{\nmp}[1]{ \nm #1 \nm_p }
\newcommand{\nmq}[1]{ \nm #1 \nm_q }

\newcommand{\IP}[2]{ \langle #1 , #2 \rangle }

\newcommand{\halmos}{\hfill $\Box$}

\newcommand{\supp}{{\rm{supp}}}

\def\Rmn{\R^{m \times n}}

\def\xh{\hat{x}}

\def\hlo{h_{\L_0}}
\def\hloc{h_{\L_0^c}}

\def\GkS{{\rm GkS}}

\def\nmsl1{\nm_{{\rm SL1}}}

\def\ct{\tilde{c}}

\def\DBP{\D_{{\rm BP}}}

\def\cb{\bar{c}}
\def\ct{\tilde{c}}
\def\Gsupp{{\rm Gsupp}}
\def\dmax{d_{{\rm max}} }
\def\dmin{d_{{\rm min}} }

{\bf}{\it}
\newtheorem{definition}{Definition}{\bf}{\it}
{\bf}{\rm}
\newtheorem{lemma}{Lemma}{\bf}{\it}
\newtheorem{theorem}{Theorem}{\bf}{\it}
{\bf}{\it}
{\bf}{\it}
{\bf}{\rm}

\begin{document}


\title{
Tight Performance Bounds for Compressed Sensing \\
With Conventional and Group Sparsity
}

\author{Shashank Ranjan and Mathukumalli Vidyasagar,
\thanks{The authors are with
the Electrical Engineering Department, Indian Institute
of Technology Hyderabad, Kandi, TS 502285, INDIA.
Emails: ee16resch11020@iith.ac.in, m.vidyasagar@iith.ac.in.
This research was supported by the National Science Foundation under
Award \#ECCS-1306630
and by the Department of Science and Technology, Government of India.
A brief description of a part of these results was presented
at the Indian Control Conference 2017 and is published
in the Proceedings \cite{MV-Ranjan17}.
}
}

\maketitle


\begin{abstract}

In this paper, we study the problem of recovering a group sparse vector
from a small number of linear measurements.
In the past the common approach has been to use various ``group
sparsity-inducing'' norms such as the Group LASSO norm for this purpose.
By using the theory of convex relaxations, we show that it is also possible
to use $\ell_1$-norm minimization for group sparse recovery.
We introduce a new concept called group robust null space property (GRNSP),
and show that, under suitable conditions, a group version of the restricted
isometry property (GRIP) implies the GRNSP, and thus
leads to group sparse recovery.
When all groups are of equal size, our bounds are less conservative than
known bounds.
Moreover, our results apply even to situations where
where the groups have different sizes.
When specialized to conventional sparsity, our bounds reduce to one of the
well-known ``best possible'' conditions for sparse recovery.
This relationship between GRNSP and GRIP is new even for conventional sparsity,
and substantially streamlines the proofs of some known results.
Using this relationship, we derive bounds on the $\ell_p$-norm of the
residual error vector for all $p \in [1,2]$, and not just when $p = 2$.
When the measurement matrix consists of random samples of a sub-Gaussian
random variable, we present bounds on the number of measurements, which
are less conservative than currently known bounds.
	
\end{abstract}

\section{Introduction}\label{sec:intro}

Compressed sensing refers to the recovery of high-dimensional vectors with
very few nonzero components from a limited number of linear measurements.
This is referred to here as the ``conventional'' sparsity problem,
and it has been the subject of a great deal of research.
In recent years, attention has also been focused on
the ``group sparsity'' problem, where there is additional information
available about the locations of the nonzero components of the unknown vector.
In this paper, we advance the status of knowledge in compressed sensing
for both conventional as well as group sparsity.
Precise details are given in subsequent sections, but in brief the
contributions of the paper are the following:
\bit
\item In conventional sparsity, the two most widely used techniques are
RIP (restricted isometry property) and the RNSP (robust null space property);
very few papers relate the two
approaches.\footnote{All terms are defined in subsequent sections}.
One of the currently best available results \cite{CZ14} on the use of the
RIP states that if 
the measurement matrix $A$ satisfies the RIP of order $tk$ for some
$t \geq 4/3$,
then it is possible to achieve robust $k$-sparse recovery via the
basis pursuit formulation, that is, minimizing an $\ell_1$-norm
objective function.\footnote{Other complementary results from
\cite{CWX10c,Zhang-Li-TIT18} are also discussed below.}
Moreover, this bound is tight, as shown in \cite{CZ14}.
In the present paper, we offer two improvements to these results.
First, we show that the above sufficient condition continues to be
sufficient whenever $t > 1$, and not just when $t \geq 4/3$.
Second, we prove this by showing that in this case the RIP implies the RNSP.
Ours is the best available relationship between RIP and RNSP.
Moreover, the connection between RIP and RNSP allows us to prove bounds
on the $\ell_p$-norm of the residual error for all $p \in [1,2]$,
and not just the $\ell_2$-norm.
The papers based on the RIP alone are not able to prove such bounds.
\item In group sparsity, until now researchers have replaced the $\ell_1$-norm
objective function by various
``group sparsity-inducing'' norms in order to achieve robust recovery.
In the present paper, we show that the standard $\ell_1$-norm can also
be interpreted as the convex relaxation of two distinct group sparsity
indices, so that $\ell_1$-norm minimization also has the potential to
achieve group sparse recovery.
Then we proceed to derive conditions under which $\ell_1$-norm minimization
actually achieves group sparse recovery.
These conditions reduce to those for conventional sparsity when all
``groups'' consist of one element each.
Our method of proof is based on the group version of the RIP, but also
a new (though very natural) group version of the RNSP.
As with conventional sparsity, we show that GRIP implies the GRNSP.
Thus, using our approach, we can derive bounds on the $\ell_p$-norm
of the residual error for all $p \in [1,2]$, which are generally not
available with group sparsity-inducing norms.
We also derive bounds on the number of samples that suffice to achieve
group sparse recovery when the measurement matrix consists of random
sub-Gaussian samples.
These bounds are smaller than currently available bounds from other papers.
Not surprisingly, it is also shown that group sparse recovery can
be achieved with fewer samples than for conventional sparse recovery.
Given that there are now very efficient methods for $\ell_1$-norm minimization,
our results suggest that $\ell_1$-norm minimization is a viable alternative
to the use of group sparsity-inducing norms, for problems of group sparse
recovery.
\eit

\section{Conventional Sparsity}\label{sec:conv}

\subsection{Summary of Some Compressed Sensing Results}\label{def:11}

Let $\SI_k \seq \R^n$ denote the set of $k$-sparse vectors in $\R^n$; that is
\bd
\SI_k := \{ x \in \R^n : \nmm{x}_0 = | \supp(x) | \leq k \} ,
\ed
where, as is customary, $\nmm{\cdot}_0$ denotes the number of nonzero
components of $x$.
Given a norm $\nmm{\cdot}$ on $\R^n$, the \textbf{$k$-sparsity index}
of $x$ with respect to that norm is defined by
\bd
\s_k(x, \nmm{\cdot}) := \min_{z \in \SI_k} \nmm{x-z} .
\ed
Now we can define the conventional compressed sensing problem precisely.

\begin{definition}\label{def:vec-rec}
Suppose $A \in \Rmn$ and $\D: \R^m \ap \R^n$.
The pair $(A,\D)$ is said to achieve \textbf{robust sparse
recovery of order $k$} and indices $p,q$ 
if there exist constants $C$ and $D$ such that,
for all $\eta \in \R^m$ with $\nmeu{\eta} \leq \e$, it is the case that
\be\label{eq:13}
\nmp{ \D (Ax + \eta) - x} \leq C \s_k( x , \nmq{\cdot} )
+ D \e , \fa x \in \R^n .
\ee
\end{definition}

Among the most popular decoder maps is $\ell_1$-norm
minimization, also known as basis pursuit.
When $y = Ax + \eta$ with $\nmeu{\eta} \leq \e$, it is defined as follows:
\be\label{eq:13a}
\DBP(y) := \argmin_{z \in \R^n} \nmm{z}_1 \st \nmeu{y - Az} \leq \e ,
\ee

There are two widely used sufficient conditions for basis pursuit to achieve
robust sparse recovery, namely the restricted
isometry property (RIP) and the robust null space property
(RNSP).
We begin by discussing the RIP.

\begin{definition}\label{def:RIP}
A matrix $A \in \Rmn$ is said to satisfy the \textbf{restricted isometry
property (RIP)} of order $k$ with constant $\d$ if
\be\label{eq:14}
(1 - \d) \nmeusq{u} \leq \nmeusq{Au} \leq (1 + \d) \nmeusq{u}  , \fa u\in\Sigma_k.
\ee
\end{definition}

Starting with \cite{Candes-Tao05}, it is shown in a series of papers that
the RIP of $A$ is sufficient for $(A,\DBP)$ to achieve robust sparse recovery.
In \cite{Candes08} it is shown that $\d_{2k} < \sqrt{2} - 1$ is sufficient
for robust $k$-sparse recovery.
This bound has been subsequently improved in several papers, but to save
space, we cite only the most recent ``best possible'' results
relating RIP and robust recovery.

\begin{theorem}\label{thm:CZ}
If $A$ satisfies the RIP of order $tk$ with constant $\d_{tk} 
< \sqrt{(t-1)/t}$ for some $t \geq 4/3$, or with constant $\d_{tk} < t/(4-t)$
for some $t \in (0,4/3)$,\footnote{Here and elsewhere. when we write 
$\d_{\al}$ and $\al$ is not necessarily an integer, we mean 
$\d_{\lceil \al \rceil}$.}
then $(A,\DBP)$ achieves robust sparse recovery with $q = 1$ and $p = 2$.
Moreover, both bounds are tight.
\end{theorem}
Note that the first bound is proved in \cite{CZ14}, while the second bound
is proved in \cite{Zhang-Li-TIT18}.
In \cite{CWX10c}, it is shown that $\d_k < 0.307$ is sufficient, which
is slightly worse than the bound $\d_k < 1/3$ implied by 
\cite{Zhang-Li-TIT18}.

An alternative to the RIP approach to compressed sensing is provided by
the robust null space property; see \cite{Foucart14} or
\cite[Definition 4.21]{FR13}.

\begin{definition}\label{def:RNSP}
A matrix $A \in \Rmn$
is said to satisfy the \textbf{$\ell_2$-robust null space property (RNSP)}
of order $k$ with constants $\r \in (0,1)$ and $\t > 0$ if, 
for every set $S \seq [n]$ with $|S| \leq k$, we have that\footnote{Note that,
for the sake of
consistency, we have introduced a factor of $\sqrt{k}$ to divide $\t$
in \eqref{eq:16}.}
\be\label{eq:16}
\nmeu{h_S} \leq \frac{\r}{\sqrt{k} } \nmm{h_{S^c}}_1 +
\frac{\t}{\sqrt{k} } \nmeu{Ah} , \fa h \in \R^n .
\ee
\end{definition}

Schwarz' inequality implies that if $A$ satisfies the $\ell_2$-RNSP,
then for every set $S \seq [n]$ with $|S| \leq k$ it also satisfies
\be\label{eq:17}
\nmm{h_S}_1 \leq \r \nmm{h_{S^c} }_1 + \t \nmeu{Ah} , \fa h \in \R^n .
\ee
This is sometimes called the RNSP without the prefix ``$\ell_2$.''

\begin{theorem}\label{thm:RNSP}
(See \cite[Theorems 4.19 and 4.22]{FR13}.)
Suppose $A$ satisfies \eqref{eq:17} with constants $\r$ and $\t$.
Then the pair $(A,\DBP)$ achieves robust $k$-sparse recovery for
$p = q = 1$, with
\be\label{eq:18}
C = 2 \frac{ 1+\r }{ 1 - \r } , D = \frac{ 4 \t}{1-\r} .
\ee
If $A$ satisfies \eqref{eq:16}, then $(A,\DBP)$ achieves robust $k$-sparse
recovery for $p = 1$ and all $q \in [1,2]$.
\end{theorem}


There are relatively few results relating the RIP and the RNSP.
Currently the best available result is \cite[Proposition 8]{Foucart14},
in which it is shown that if $A$ satisfies the RIP of order $2k$
with constant $\d_{2k} < 1/9$, then it also satisfies the RNSP.
Note that $1/9$ is far smaller than $\sqrt{1/2}$ which is the bound
on $\d_{2k}$ from Theorem \ref{thm:CZ}.

\subsection{Our Contributions}\label{ssec:12}

Against this background, in this paper
we show that, if $A$ satisfies the RIP of order $tk$ with constant
$\d_{tk} < \sqrt{(t-1)/t}$ for some $t > 1$, then
$A$ also satisfies the $\ell_2$-RNSP for appropriate constants;
see Theorem \ref{thm:1a}.
This has several consequences.
First, this is by far the best result that relates RIP to RNSP.
Second, the limit on $t$ is reduced from $t \geq 4/3$ in Theorem \ref{thm:CZ}
to $t > 1$.
Third, by establishing that the condition $\d < \sqrt{(t-1)/t}$ implies
the $\ell_2$-RNSP, we can establish that for such a matrix $A$,
basis pursuit achieves robust $k$-sparse recovery for all $p \in [1,2]$
(and $q = 1$), and are able to prove bounds on the $\ell_p$-norm
of the residual for all $p \in [1,2]$.
This is in contrast to most existing papers including \cite{CZ14} where
robust $k$-sparse recovery is established using the RIP, and thus
only for $p = 2$.
Moreover, our bounds are an improvement over those in 
\cite[Theorem 4.25]{FR13}.

\section{Group Sparsity}\label{sec:group}

\subsection{Literature Review}\label{ssec:21}

At about the same time that the problem of robust sparse recovery was
being addressed via $\ell_1$-norm minimization, the research community
began to propose that the number of nonzero components of a vecor might not
be the only reasonable measure of the sparsity of a vector.
Alternate notions under the broad umbrella of ``group sparsity'' 
and ``group sparse recovery'' began to
appear, starting with \cite{Yuan-Lin-JRSSB06}.
In its simplest form, group sparsity refers to the case where the index set
$[n]$ is partitioned into $g$ disjoint sets $G_1 , \ldots , G_g$.
In the early papers such as 
\cite{Stojnic-et-al-TSP09,Baraniuk-et-al-TIT10,Eldar-Mishali-TIT09,Eldar-et-al-TSP10}, it is assumed that all groups $G_i$ have the same size $d$,
so that $n = gd$.
However, starting with \cite{Huang-Zhang-AS10}, the groups are not
required to have a common size.
In almost all current papers on group sparse recovery,
the $\ell_1$-norm objective function in \eqref{eq:13a} is changed 
to the so-called \textbf{Group LASSO norm}
introduced in \cite{Yuan-Lin-JRSSB06}, defined as
\be\label{eq:23}
\nmm{x}_{{\rm GL}} := \sum_{j \in [g]} \nmeu{x_{G_j} } ,
\ee
where $x_{G_j}$ denotes the projection of $x$ onto the set $G_j$.

For this formulation,
a variety of recovery results are proved by several authors.
In \cite{Eldar-Mishali-TIT09}, a block RIP analogous to the RIP
is introduced, and it is shown that recovery of group $k$-sparse 
vectors\footnote{Please see Section for the definition.} is achieved
by minimizing the GL norm $\nmm{x}_{{\rm GL}}$
if $\d_{B,2k} < \sqrt{2} - 1$, which is an extension of the conventional
sparsity result in \cite{Candes08}.
In \cite{Eldar-et-al-TSP10}, a notion of block-coherence is introduced;
it is shown that, just as in conventional sparsity, block-coherence
implies block-RIP, which in turn implies group sparse recovery.
The well-known orthogonal matching pursuit algorithm \cite{PRK-OMP}
is modified to a block-OMP and it is shown that block-OMP recovers
a block-sparse signal under suitable conditions.
In \cite{Stojnic-et-al-TSP09,Baraniuk-et-al-TIT10,Huang-Zhang-AS10},
the measurement matrix $A$ is assumed to be a randomly generated Gaussian
matrix, and bounds are derived on the number of samples that
sufficie for the (probabilistic) recovery of
a group-sparse vector $x$ by minimizing the GL norm.
In the first two papers, it is assumed that all groups have the same size
and it is shown that the required number of measurements with group sparsity
is less than with conventional sparsity.
In \cite{Baraniuk-et-al-TIT10}, the behavior of well-known algorithms
such as CoSaMP and IHT (iterative hard thresholding) is analyzed with the
Group LASSO norm.
In \cite{Huang-Zhang-AS10} the authors dispense with the requirement of
equal group sizes, and derive a very restrictive sufficient condition 
(see \cite[Assumption 4.3]{Huang-Zhang-AS10}) for group
sparse recovery.
Indeed, in \cite{Huang-Zhang-AS10} the authors state ``Note that
this assumption does not show the benefit of group Lasso over standard Lasso.''
In \cite{Elham-Vidal-CVPR11}, group sizes need not be equal, and
the GL norm is modified by replacing $\nmeu{x_{G_j} }$ with $\nmq{x_{G_j} }$
for any $q \geq 1$.
Sufficient conditions for group sparse recovery are established in terms
of the group RIP, and block coherence.
This work is extended in \cite{Elham-Vidal-TSP12} to incorporate
subspace coherence, whose value is in general smaller than block coherence.

The above papers can be thought as representing the first phase of research into
group sparse recovery.
Subsequent papers follow several different and unrelated directions.
Several papers in the statistics community analyze the \textit{asymptotic}
behavior of minimizing the Group LASSO norm as $\nai$, with $k$ either
kept fixed, or increasing more slowly than $n$.
Some of these papers also study the problem of ``simultaneous'' estimation
of several unknown vectors that share a common sparsity pattern, using
a common measurement matrix.
In \cite[Corollary 4.1, Theorem 7.1]{Lounici-et-al-AS11}, it is shown
that \textit{in the problem of simultaneous estimation}, the Group LASSO norm
offers advantages over the standard $\ell_1$-norm.
In \cite{Obozinski-et-al-AS11}, the authors study the problem of
\textit{support} recovery, that is, recovering the set of nonzero
components of the unknown vector, under group sparsity.
They give a very tight bound on the rate at which the number of samples
must grow in order to achieve support recovery.
These results show that, when the unknown vector is supported over
\textit{a union} of unknown subspaces, the Group LASSO formuation is natural.
Support recovery is also the subject of \cite{Elyad-et-al-ICASSP17}.
Unlike other papers, the results in this paper are not asymptotic.
vector itself.
Note that it is possible to recover the support of an unknown vector while not
recovering the vector itself.
On the other hand, if the nondominant components of a (group) sparse
vector are much smaller than the dominant ones, recovering a good
approximation to the unknown vector also leads to support recovery.

In \cite{Rao-et-al-AI-Stat12} and the references therein,
the emphasis is on removing the assumption that the sets $G_j$ are
pairwise disjoint; thus the focus is on \textit{overlapping} group
decompositions.
In \cite{Yang-Zhang-Xie-TSP12,Tan-Yang-Nehorai-TSP14}, the authors study the
case where there is uncertainty and/or error in implementing the measurement
matrix $A$.
Instead of the designed matrix $A$, the measurements equal $y = (A + B E)x$
for suitable models of $B,E$.
One of the important innovations of these papers is the notion of ``joint''
sparsity.
To illustrate, suppose $n = 2l$.
Then for a given $k < n$, a vector $x$ is said to be \textbf{jointly $k$-sparse}
if its support is concentrated a set of the form $S \cup (l+S)$,
where $|S| \leq k$, and $l+S$ denotes shifting every element of $S$ by $l$.
This model is apparently very natural in problems of detecting the Direction
of Arrival (DoA).
A joint RIP is defined for such vectors.
It is clear that the joint $2k$ RIP constant is smaller than not just
the standard $2k$ RIP constant, but also the group $2k$
RIP constant, because of the restrictions on the support set.
Therefore, group sparse recovery would require fewer samples than 
conventional sparse recovery, while joint sparse recovery would require
still fewer samples.
Finally, in \cite{Bajwa-et-al-TIT15}, the authors relax the requirement
from recovering \textit{every} group sparse vector to \textit{average case}
recovery.
Naturally, the sufficient conditions for recovery in this case are weaker
than for the recovery of \textit{every} vector.
The main drawback of this approach is that there is no way to know whether
the \textit{particular} group sparse vector that one is attempting to
recover lies within the set of recoverable vectors.

The above discussion can be briefly summarized as follows:
The Group LASSO formulation is better than the conventional LASSO formulation
when it comes to simultaneous estimation, and in support recovery.
In the present paper, the focus is on estimating a single vector (therefore
not simultaneous recovery, nor support recovery).
In the opinion of the authors, currently available results such as
{Lounici-et-al-AS11} do not establish conclusively whether Group LASSO
offers any unambiguous advantages in this situation.

\subsection{Basis Pursuit as a Group Sparsity-Inducing Norm}\label{ssec:22}

In the conventional setting, the quantity $\nmm{x}_0$ which counts the
number of nonzero components of $x$ is taken as a measure of the sparsity of
$x$.
In the case of group sparsity, it is possible to think of two distinct-looking
definitions.\footnote{We thank one of the reviewers for suggesting this
notation.}
\be\label{eq:21}
\nmm{x}_{{\rm UG,0}} = \sum_{j \in [g]} \oneb_{ \{ x_{G_j} \neq \bz \} } ,
\ee
\be\label{eq:22}
\nmm{x}_{{\rm G,0}} = \sum_{j \in [g]} |G_j| \oneb_{ \{ x_{G_j} \neq \bz \} } ,
\ee
where $x_{G_j}$ denotes the projection of $x \in \R^n$ onto the indices
in $G_j$, and $\oneb$ denotes the indicator function.
Thus $\nmm{x}_{{\rm UG,0}}$ counts the number of groups on which $x$
has a nonzero projection, whereas $\nmm{x}_{{\rm G,0}}$ counts the
cardinality of the union of groups over which $x$ has a nonzero projection.
It is obvious that if all groups have the same size $d$ (as was assumed in many
early papers), then both definitions differ only by a factor of $d$.
However, when group sizes differ widely, the two quantities can be very
different.
While a majority of papers use the definition in \eqref{eq:21},
\cite{Huang-Zhang-AS10} uses a combination of both parameters.

Let us define a vector $x \in \R^n$ to be \textbf{$l$-group sparse} if
$\nmm{x}_{{\rm UG,0}} \leq l$, and \textbf{group $k$-sparse} if
$\nmm{x}_{{\rm G,0}} \leq k$.
Further, define $\dmax$ and $\dmin$ denote the largest and smallest
group sizes.
Then an $l$-group sparse vector is also $l \dmax$-sparse in the conventional
sense, but the converse is not true.
Similarly, a group $k$-sparse vector is also $k$-sparse in the conventional
sense, but the converse is not true.
In the proofs of Theorems \ref{thm:1} and \ref{thm:1b}, we make use of the
fact that there is a known prior bound on the sparsity count of the
unknown vector, irrespective of the number of groups over which it is
supported.
Thus we prefer to work with group $k$-sparse vectors and not
$l$-group sparse vectors.
In principle our proofs could be adapted to $l$-group sparse vectors by
treating them as group $l \dmax$-sparse vectors.
Working with the latter would lead to more conservative bounds for recovery.

In the case of conventional sparsity, replacing the nonconvex objective function
$\nmm{x}_0$ by $\nmm{x}_1$ is justified using the concept of a convex relaxation.
It is now shown that \textit{every}
group decomposable norm on $\R^n$, \textit{including} the $\ell_1$-norm,
is the convex relaxation of both $\nmm{\cdot}_{{\rm UG,0}}$ 
and $\nmm{\cdot}_{{\rm G,0}}$ over suitably defined convex sets.

Suppose $\OM \seq \R^n$ is a convex set, and that $f: \OM \ap \R$.
Then a function $g: \OM \ap \R$ is said to be the \textbf{convex relaxation}
of $f$ over $\OM$ if: (i) $g(x) \leq f(x) \fa x \in \OM$, and (ii)
if $h: \OM \ap \R$ is convex and satisfies $h(x) \leq f(x) \fa x \in \OM$,
then $h(x) \leq g(x) \fa x \in \OM$.
In other words, the convex relaxation of $f$ is the largest convex
function that is dominated by $f$ on the set $\OM$.
Observe that the same function $f$ but on a different convex set $\OM'$
could have a different convex relaxation $g'$.
There is a conceptually simple way to determine the convex relaxation, namely
through the use of convex duality.
Theorem \cite[Theorem E.1.3.5]{H-UL01} states that the second dual of $f$
is its convex relaxation.
Moreover, using the definition of the dual, it is easy to establish
the following fact.

\begin{lemma}\label{lemma:dual1}
Let $\{ G_1 , \ldots , G_j \}$ be a partition of $[n]$.
Write $\R^n = \prod_{j=1}^g \R^{|G_j|}$, and suppose that
$\OM \seq \R^n = \prod_{j=1}^g \OM_j$ where each $\OM_j \seq \R^{|G_j|}$.
Further, suppose $f: \OM \ap \R$ is decomposable as
\be\label{eq:28}
f(x) = \sum_{j=1}^g f_j(x_{G_j}) ,
\ee
where $x_{G_j}$ is the projection of $x$ onto $\R^{|G_j|}$.
Then the convex relaxation $g$ of $f$ equals
\be\label{eq:29}
g(x) = \sum_{j=1}^g g_j(x_{G_j}) ,
\ee
where $g_j$ is the convex relaxation of $f_j$ over $\OM_j$.
\end{lemma}

The next lemma is also easy to prove.

\begin{lemma}\label{lemma:dual2}
Suppose $c > 0$ is some constant, and let $c \cdot \B$
denote the set of all $x \in \R^l$ with $\nmm{x} \leq c$.
Then the convex relaxation of $\phi$ over $c \cdot \B$ is $(1/c) \nmm{\cdot}$.
\end{lemma}

Next, let us refer to a norm $\nmm{\cdot}$ on $\R^n$ as \textbf{group
decomposable} if it is of the form
\be\label{eq:210}
\nmm{x} = \sum_{j \in [g]} \nmm{x_{G_j} }_{G_j} ,
\ee
for suitably defined norms $\nmm{\cdot}_{G_j}$ on $\R^{|G_j|}$.
The next result follows as a ready consequence of Lemmas \ref{lemma:dual1}
and \ref{lemma:dual2}.

\begin{lemma}\label{lemma:dual3}
Let $\nmm{\cdot}_{G_j}, j \in [g]$ be arbitrary norms on $\R^{|G_j|}$.
Let $\OM_j \seq \R^{|G_j|}$ denote the unit ball of $\nmm{\cdot}_{G_j}$,
and define
$\OM = \prod_{j \in [g]} \OM_j$.
Then the convex relaxation of $\nmm{\cdot}_{{\rm UG,0}}$ over $\OM$ 
is the norm defined in \eqref{eq:210}.
More generally, define $\OM'_j = |G_j| \cdot \OM_j$ for all $j \in [g]$,
and let $\OM' = \prod_{j \in [g]} \OM'_j$.
Then the convex relaxation of $\nmm{\cdot}_{{\rm G,0}}$ over $\OM'$ 
is the norm defined in \eqref{eq:210}.
\end{lemma}

In short, \textit{every} group decomposable norm is the convex relaxation of
$\nmm{\cdot}_{{\rm UG,0}}$ over a suitably defined product set $\OM$.
Conversely, the convex relaxation of $\nmm{\cdot}_{{\rm UG,0}}$
over every product set is a group decomposable norm.
Moreover, every convex relaxation of $\nmm{\cdot}_{{\rm UG,0}}$ over
a product $\OM$ is also a convex relaxation of $\nmm{\cdot}_{{\rm G,0}}$
over the related set $\OM'$, and vice versa.
In particular, if we were to choose each of the sets $\OM_j$
to be the unit balls in the $\ell_2$-norm over the corresponding
space, then the convex relaxation of $\nmm{\cdot}_{{\rm UG,0}}$ over $\OM$
would be the Group LASSO norm defined in \eqref{eq:23}.
However, if we were to choose each of the sets $\OM_j$
to be the unit balls in the $\ell_1$-norm over the corresponding space,
then the convex relaxation of $\nmm{\cdot}_{{\rm UG,0}}$ over $\OM$ would be the
$\ell_1$-norm!
The point is that, in principle, even basis pursuit can be used to
achieve group sparse recovery, even though it is not obviously
``group-sparsity inducing.''

\subsection{Our Contributions}\label{ssec:23}

In the present paper, we use $\ell_1$-norm minimization,
and establish that group sparse recovery results under appropriately
defined sufficient conditions. 
These are based on group analogs of the RIP and the RNSP
for group $k$-sparse vectors.
So far as we are able to determine, this is the first time that a
group version of the RNSP is proposed and used to establish group
sparse recovery.
The bounds derived here
are less conservative than those proved earlier by others, based on the
group RIP, both for the
case where all groups are of equal size
\cite{Eldar-Mishali-TIT09,Eldar-et-al-TSP10,Baraniuk-et-al-TIT10},
and are of unequal size
\cite{Elham-Vidal-CVPR11,Elham-Vidal-TSP12}.
When the measurement matrices consist of random samples of
sub-Gaussian variables, and all groups are of equal size, our estimates
are of the same order as in
\cite{Baraniuk-et-al-TIT10}
and are smaller than for conventional sparse recovery.
When group sizes are unequal and the measurement matrix is random,
our bounds are far less conservative than those in 
\cite[Assumption 4.3, Theorem 5.1]{Huang-Zhang-AS10}.
It is of course possible results similar to ours could be established
using the Group LASSO norm instead of the $\ell_1$-norm.
That would be a topic for future research.

The rest of the paper is organised as follows:
The main results of the paper concerning group sparse recovery
and concerning conventional sparse recovery are stated in
Sections \ref{sec:new-1} and  \ref{sec:new-2} respectively.
These results are compared against known results in Section \ref{sec:disc}.
Numerical examples are given in Section \ref{sec:exam}, and
the proofs of the main results 
are given separately in Section \ref{sec:proofs}.
Throughout the paper, we use the basis pursuit denoising approach.
Therefore, given $y = Ax + \eta$ with $\nmeu{\eta} \leq \e$, we define
\be\label{eq:11a}
\xh = \argmin_z \nmm{z}_1 \st \nmeu{y - Az} \leq \e .
\ee

\section{Main Results - I: Group Sparse Recovery}\label{sec:new-1}

We say that a vector $u$ is \textbf{group $k$-sparse} if 
$\nmm{u}_{{\rm G,0}} \leq k$, where $\nmm{\cdot}_{{\rm G,0}}$ is defined
in \eqref{eq:22}.
We also require the notion of a group $k$-sparse subset of the index set $[n]$.
Recall that $\{ G_1 , \ldots , G_g \}$ is a partition of $[n]$.
If $L \seq [g]$, let $G_L$ denote $\cup_{j \in L} G_j$.
Then a set $S \seq [n]$ is said to be a \textbf{group $k$-sparse subset}
of $[n]$ if $S = G_L$ for some subset $L \seq [g]$, and moreover, $|S| \leq k$.
Note that a vector is a group $k$-sparse vector if and only if 
$\supp(x)$ is a group $k$-sparse set.
We denote the set of all group $k$-sparse vectors by $\SI_{G,k}$,
and the collection of all group $k$-sparse sets by $\GkS$.

We begin by defining group analogs of the RIP and RNSP for
group $k$-sparse vectors.

\begin{definition}\label{def:GRIP}
A matrix $A \in \R^{m \times n}$ is said to satisfy the \textbf{group
restricted isometry property (GRIP)}
of order $k$ with constant $\d_{G,k} \in (0,1)$ if
\be\label{eq:31a}
(1 - \d_{G,k} ) \nmeu{u} \leq \nmeu{Au} \leq (1 + \d_{G,k}) \nmeusq{Au},
\fa u \in \SI_{G,k} .
\ee
\end{definition}

\begin{definition}\label{def:GRNSP}
A matrix $A \in \R^{m \times n}$ is said to satisfy the
\textbf{$\ell_2$-group robust null space property (GRNSP)}
with constants $\r_G \in (0,1), \t_G \in \R_+$,
if, for all $h \in \R^n$ and all sets $S \in \GkS$, it is true that
\be\label{eq:314}
\nmeu{h_{S}} \leq \frac{\r_G}{\sqrt{k}} \nmm{ h_{S^c} }_1
+ \frac{\t_G}{\sqrt{k}} \nmeu{Ah} .
\ee
\end{definition}

As with RNSP,
Schwarz' inequality implies that if $A$ satisfies the $\ell_2$-GRNSP, then
for all $h \in \R^n$ and all sets $S \in \GkS$, it is true that
\be\label{eq:314a}
\nmm{ h_S }_1 \leq \r_G \nmm{ h_{S^c} }_1 + \t_G \nmeu{ Ah }.
\ee

\subsection{Group Robust Null Space Property}\label{ssec:GRNSP}

Now we present the first of our main results, which
allows us to establish robust group $k$-sparse recovery.
For notational convenience, define
\bd
\dmax = \max_{j \in [g]} |G_j| , \dmin = \min_{j \in [g]} |G_j| ,
\ed
Given integers $k, n$ and a real number $t > 1$, define
\be\label{eq:311d}
\kb := [1 + (t-1) \dmax] k .
\ee
Also define
\be\label{eq:310}
\nu:=\sqrt{(t-1)t}-(t-1) .
\ee
It is easy to verify via elementary calculus that $\nu = 0$ if $t = 1$,
$\nu$ is an increasing function of $t$, and $\nu \ap 0.5$
as $\tai$.

\begin{theorem}\label{thm:1}
Suppose that the matrix $A$ satisfies the GRIP of order
$\kb$
with constant $\d_{G,\kb} < \bar{\d}_G$, where\footnote{As before,
when we write
$\d_{G,\al}$ and $\al$ is not necessarily an integer, we mean
$\d_{G,\lceil \al \rceil}$.}
\be\label{eq:315}
\bar{\d}_G = \nu (1 - \nu) \left( \frac{ \nu^2 \dmax}{2(t-1) \dmin} +
0.5 - \nu + \nu^2 \right)^{-1} .
\ee
Then $A$ satisfies the $\ell_2$ GRNSP with constants $\r_G, \t_G$ defined as
follows:
\be\label{eq:313}
\r_G := c_G/a < 1 , \t_G := b\sqrt{k}/a^2 ,
\ee
where
\beq
a & := & [\nu(1-\nu)-\delta(0.5-\nu(1 - \nu))]^{1/2} \nonumber \\
& = & \frac { [ (1-\d)-(1+\d)(1-2\nu)^2 ]^{1/2} }{2}  , \label{eq:311}
\eeq
\be\label{eq:312}
b :=\nu(1-\nu)\sqrt{1+\delta} ,
\ee
\be\label{eq:312a}
c_G := \left[ \frac{ \d \nu^2 \dmax } {2(t-1)\dmin } \right]^{1/2} ,
\ee
and $\d$ is shorthand for $\d_{G,\kb}$.
\end{theorem}

A simplification is possible in the case where
all groups have the same size, so that $\dmax = \dmin$.

\begin{theorem}\label{thm:1b}
Suppose $n = dg$ for some integers $d,g$, and that all groups have size $d$.
Suppose the matrix $A$ satisfies the GRIP of order $\kb$ with constant
$\d_{G,\kb} < \bar{\d} = \sqrt{(t-1)/t}$.
Then $A$ satisfies the $\ell_2$ GRNSP with constants $\r, \t$ defined as
follows:
\be\label{eq:313e}
\rho := c/a < 1 , \tau := b\sqrt{k}/a^2 ,
\ee
where $a, b$ are as in \eqref{eq:311} and \eqref{eq:312} respectively, and
\be\label{eq:312b}
c:= \left[ \frac{ \d \nu^2  } {2(t-1) } \right]^{1/2} ,
\ee
and $\d$ is shorthand for $\d_{G,\kb}$.
\end{theorem}

\subsection{Error Bounds on the Residual Vector}\label{ssec:bounds}

Suppose $x$ is the unknown vector and $\xh$ is the recovered vector,
constructed according to \eqref{eq:11a}.
In this subsection we present bounds for
$\nmp{\xh - x}$ for $p \in [1,2]$.

\begin{theorem}\label{thm:2}
Suppose that the measurement matrix $A$ satisfies the conditions of
Theorem \ref{thm:1}.
Then the formulation \eqref{eq:11a} achieves robust group sparse
recovery of order $k$.
Specifically, let $\xh$ be defined as in \eqref{eq:11a}, and let $h = \xh - x$
denote the residual error vector.
Then
\begin{equation}\label{eq:350}
\nmm{h}_1\leq\ \frac{2}{1-\r_G}[(1+\r_G) \s_{G,k}(x,\nmm{\cdot}_1)
+\ 2\t_G\epsilon] ,
\end{equation}
and for all $p \in [1,2]$,
\beq
\nmp{h} & \leq & \frac{2}{1-\r_G} \left( 1 + \frac{\r_G}{k^{1-1/p}}
\right) \s_{G,k}(x,\nmm{\cdot}_1)
\nonumber \\ 
& + & \left[ \frac{2}{1-\r_G} \left( 1 + \frac{\r_G}{k^{1-1/p}} \right)
+ \frac{2}{k^{1-1/p}} \right] \t_G \e , \label{eq:350a}
\eeq
where both $\r_G$ and $\t_G$ are defined in \eqref{eq:313},
and $\s_{G,k}(x,\nmm{\cdot}_1)$ denotes the group $k$-sparsity index of $x$
defined by
\bd
\s_{G,k}(x,\nmm{\cdot}_1) = \inf_{S \in \GkS} \nmm{x - x_S}_1 .
\ed
\end{theorem}

\subsection{Sample Complexity Estimates}\label{ssec:sample}

In this subsection we study the case where the measurement matrix $A$
equals 
\be\label{eq:350b}
A = (1/\sqrt{m}) \Phi ,
\ee
where $\Phi$ consists of $mn$ independent
samples of a zero-mean, unit-variance random variable $X$ that satisfies
%
\be\label{eq:28b}
E[ \exp(\th X)] \leq \exp(\cb \th^2), \fa \th \in \R .
\ee
for some constant $\cb$.
Such a random variable is said to be \textbf{sub-Gaussian}.
In such a case, it can be shown that there exists
a constant $\ct$ such that
\be\label{eq:28c}
\Pr \{ | \nmeusq{Au} - \nmeusq{u} | \geq t \nmeusq{u} \}
\leq 2 \exp( \ct m t^2) , \fa t \in \R .
\ee
The relationship between the sub-Gaussian parameter $\cb$
in \eqref{eq:28b} and the constant $\ct$ can be derived by combining
various arguments in \cite{FR13}.
See in particular \cite[Lemma 9.8]{FR13}.

\begin{lemma}\label{lemma:21}
Suppose $X$ is a zero-mean, unit variance random variable, and satisfies
\eqref{eq:28b} for some constant $\cb$.
Define
\be\label{eq:28d}
\g = 2 , \zeta = 1/(4\cb), \al = \g e^{-\zeta} + e^\zeta .
\ee
Then \eqref{eq:28c} is satisfied with
\be\label{eq:28e}
\ct = \frac{\zeta^2}{2( 2 \al + \zeta) } .
\ee
\end{lemma}

By adapting \cite[Theorem 9.11]{FR13} via replacing $k$ by $tk$ throughout,
we can give a bound on the number of measurements $m$ that suffice to
ensure that $A$ defined in \eqref{eq:350b}
satisfies the RIP or order $tk$ with constant $\d_{tk} < \d$,
with probability $\geq 1 - \xi$.

%

\begin{theorem}\label{thm:27}
Given integers $n, k < n$ and a small number $\xi \in (0,1)$, choose
any $t > 1$ and any $\d < \sqrt{(t-1)/t}$.
Let $X$ be a sub-Gaussian random variable that satisfies \eqref{eq:28b}
for some $\cb > 0$, and define $A$ as in \eqref{eq:350b}.
Choose an integer $m_C$ such that
\be\label{eq:28g}
m_C \geq \frac{1}{ \ct \d^2 } \left( \frac{4}{3} tk \ln \frac{en}{tk}
+ \frac{14tk}{3} + \frac{4}{3} \ln \frac{2}{\xi} \right) .
\ee
Then $A$ satisfies the RIP of order $tk$ with constant $\d_{tk} < \d$
with probability $\geq 1 - \xi$.
Consequently the
pair $(A,\DBP)$ achieves robust sparse recovery of order $k$
with probability at least $1 - \xi$.
\end{theorem}

Now we present our first original result, which is an extension of
Theorem \ref{thm:27} to group sparse recovery.

\begin{theorem}\label{thm:28}
Given integers $n, k$, choose any $\d < \bar{\d}_G$ where $\bar{\d}_G$
is defined in \eqref{eq:315}.
Choose any $t > 1$, define $\kb$ as in \eqref{eq:311d}, and define
\be\label{eq:41a}
\phi = \left\lceil  \frac{ \kb }{ \dmin } \right\rceil .
\ee 
Let $X$ be a sub-Gaussian random variable that satisfies \eqref{eq:28b}
for some $\cb > 0$, and define $A$ as in \eqref{eq:350b}.
Choose an integer $m_G$ such that
\be\label{eq:41}
m_G \geq \frac{1}{ \ct \d^2 } \left( \frac{4}{3} \phi \ln \frac{eg}{\phi}
+ \frac{14\phi}{3} + \frac{4}{3} \ln \frac{2}{\xi} \right) .
\ee 
Then $A$ satisfies the GRIP of order $\kb$ with constant $\d_{\kb} < \d$
with probability $\geq 1 - \xi$.
Consequently the pair $(A,\DBP)$ achieves robust group sparse recovery of
order $k$ with probability at least $1 - \xi$.
\end{theorem}



Now let us specialize Theorem \ref{thm:28} to the case where all groups
have the same size $d$ (so that $n = gd$), and in addition, $k = ld$
for some integer $l$.
In this case \eqref{eq:41a} becomes
\bd
\phi = [1 + (t-1) d] l ,
\ed
while the bound for $m_G$ in \eqref{eq:41} becomes
\be\label{eq:43}
m_G \geq \frac{1}{ \ct \d^2 } \left( \frac{4}{3} \phi \ln \frac{eg}{\phi}
+ \frac{14\phi}{3} + \frac{4}{3} \ln \frac{2}{\xi} \right) .
\ee
The above bound $m_G$ for the number of samples that suffices
for robust \textit{group} sparse recovery should be compared
against the number $m_C$ for conventional sparsity in \eqref{eq:28g}.
In this case the estimate for $m_C$ from \eqref{eq:28g} becomes
\be\label{eq:42}
m_C \geq \frac{1}{ \ct \d^2 } \left( \frac{4}{3} tld \ln \frac{eg}{tl}
+ \frac{14tld}{3} + \frac{4}{3} \ln \frac{2}{\xi} \right) ,
\ee
after noting that $n/k = g/l$.

\begin{theorem}\label{thm:29}
If $d > 1$ and $tld = tk < g$, then $m_G < m_C$, where $m_G$ is defined
in \eqref{eq:43}, and $m_C$ is defined in \eqref{eq:42}.
\end{theorem}

Thus, in the case where all groups are of equal size, achieving robust
group $k$-sparse recovery requires \textit{fewer measurements}
than for conventional
sparsity, whenever $tk$ is smaller than the number of groups,
which is a very reasonable assumption.
On the other hand, if there is a very large disparity between group sizes,
the estimate given by \eqref{eq:41} could be larger than the estimate
for conventional sparsity given in \eqref{eq:28g};
however, it can also be smaller.
This is illustrated in the numerical example in Section \ref{sec:exam}.

\section{Main Results -- II: Conventional Sparse Recovery}\label{sec:new-2}

In this section we present our results regarding conventional sparsity.
The sufficient condition for sparse recovery is an immediate special case
of Theorem \ref{thm:1}.
However, the bounds for the $\ell_p$-norm of the residual error
require a separate proof.

In the case of conventional sparsity, all groups have cardinality one,
GRIP becomes RIP, and GRNSP becomes RNSP.
Thus Theorem \ref{thm:1} immediately implies the following.

\begin{theorem}\label{thm:1a}
Given integers $k, n$ and a real number $t > 1$, 
suppose that the matrix $A$ satisfies the RIP of order $tk$ with constant
$\d_{tk} = \d < \bar{\d} := \sqrt{(t-1)/t}$.
Then $A$ satisfies the RNSP with constants
\bd
\r = c/a < 1 , \t = b\sqrt{k}/a^2 ,
\ed
where $a,b,c$ are as in \eqref{eq:311}, \eqref{eq:312}, and \eqref{eq:312b}
respectively.
\end{theorem}

Because conventional sparsity is a special case of group sparsity where
each group has cardinality one, it is possible to obtain bounds from
Theorem \ref{thm:2}
to generate bounds on the residual error for conventional sparsity.
However, we can do better than this.

\begin{theorem}\label{thm:26}
Suppose that $A \in \Rmn$ satisfies the $\ell_2$-robust
null space property of order $k$
as defined in Definition \ref{def:RNSP}, and let $\xh$ denote the solution
of \eqref{eq:11a}.
Then
\be\label{eq:2217}
\nmm{ \xh - x }_1 \leq 2 \frac{1+\r}{1-\r} \s_k(x,\nmm{\cdot}_1)
+ \frac{4 \t}{1-\r} \e .
\ee
Moreover, for all $p \in [1,2]$, we have that
\be\label{eq:2226}
\nmp{ \xh - x } \leq \frac{1}{ k^{1 - 1/p} } \cdot \frac{2}{1 - \r}
[ (1+2 \r ) \s_k(x,\nmm{\cdot}_1) + 3 \t \e ] .
\ee
\end{theorem}

\section{Discussion of Our Contributions}\label{sec:disc}

\subsection{Group Sparsity}\label{ssec:61}

When all groups have the same size, the GRIP defined here is essentially
the same as the group- or block-RIP defined in earlier papers.
However, the sufficient conditions we derive are weaker.
When all groups have equal size $d$, the sufficient condition
proved here in Theorem \ref{thm:1b} is that, for some $t > 1$, we have 
\bd
\d_{G,[1+(t-1)d]k} < \sqrt{ \frac{t-1}{t} } .
\ed
In particular, if we set $t = 2$, we get the bound
\bd
\d_{G,(1+d)k} < \sqrt{1/2} \approx 0.707 .
\ed
This can be compared to the bound derived in various papers including
\cite[Theorem 1]{Eldar-Mishali-TIT09}
or \cite[Definition 2 and Theorem 1]{Elham-Vidal-CVPR11}, namely
\bd
\d_{G,2dk} < \sqrt{2} - 1 \approx 0.414 .
\ed
Obviously $\d_{G,(1+d)k} \leq \d_{G,2dk}$, and $\sqrt{2} - 1 < \sqrt{1/2}$.
Thus the bound derived here is less conservative.
Other papers that use the GRIP approach do not work with unequal
group sizes, whereas we can handle even this case.
Finally, because we prove our results by establishing the $\ell_2$-GRNSP,
we can derive bounds on the $\ell_p$-norm of the residual error for
all $p \in [1,2]$, as opposed to just the Euclidean norm in existing papers.

Next we discuss the case where the measurement matrix $A$ consists of
random samples of sub-Gaussian variables.
When all groups have the same size $d$, and $n = gd, k = ld$ for some
integers $g,l$, the number of samples becomes $O(l \log(g/l))$
as opposed to $O(k \log(n/k))$ for conventional sparsity.
This is not a novel observation, and is contained in practically every
paper in the area, e.g.\
\cite{Stojnic-et-al-TSP09,Huang-Zhang-AS10,Baraniuk-et-al-TIT10} and others.
For the
case of unequal group sizes the condition in 
\cite[Assumption 4.3]{Huang-Zhang-AS10} is (in the present
notation) a bound on
\bd
\frac{\d_{G,k+\dmax} + \d_{G,2k+2\dmax} }{ 1 - \d_{G,k+\dmax} } .
\ed
See \cite[Theorem 5.1]{Huang-Zhang-AS10}.
Clearly the bounds derived here are less conservative.
Other papers on the topic cannot handle the case where group sizes
are unequal.
Thus replacing the ``sparsity-inducing'' Group LASSO norm with the
$\ell_1$-norm apparently does \textit{not} lead to more conservative estimates
for the number of measurements.

\subsection{Conventional Sparsity}\label{ssec:62}

Note that the bound in Theorem \ref{thm:1a} is precisely the bound
given by \cite{CZ14} and stated here as Theorem \ref{thm:CZ}, but
with the restriction that $t \geq 4/3$.
Here the bound is lowered to $t > 1$.
However, the bound $\d_{tk} < t/(4-t)$ for some $t \in (0,4/3)$ proved
in \cite{Zhang-Li-TIT18} is not covered by our approach.
Moreover, the method of proof given in \cite{CZ14} does not establish
the robust null space property.
Rather, the proof is based on directly manipulating various inequalities.
Consequently in \cite{CZ14} the case of noise-free measurements and noisy
measurements are treated separately.
In addition, the proof in \cite{CZ14} leads only to a bound on the
Euclidean norm of the residual error $\xh - x$ when $\xh$ is computed via
\eqref{eq:11a}
In contrast, by first establishing that the RIP implies the RNSP, we are able
to treat the cases of noise-free and noisy measurements in a common framework,
and also
to obtain bounds on $\nmp{\xh - x}$ for all $p \in [1,2]$, and not just
for $p = 2$.

The result in Theorem \ref{thm:1a} is the best available to date
showing that RIP implies the RNSP.
Previously the best available result was \cite[Proposition 8]{Foucart14},
in which it is shown that if $A$ satisfies the RIP of order $2k$
with constant $\d_{2k} < 1/9$, then it also satisfies the $\ell_2$-RNSP.
The bound $1/9$ is far smaller than the bound $\sqrt{1/2} \approx 0.7071$
that results from Theorem \ref{thm:1a}.

Equation \eqref{eq:2217} is the same as \cite[Theorem 4.19]{FR13}.
However, our method of proof is different, and this leads to an
improvement in the bounds for the $\ell_p$-norm of the residual error
when $p > 1$, when compared to \cite[Theorem 4.25]{FR13}.
The bound in \eqref{eq:2226} is an improvement over that in
\cite[Theorem 4.25]{FR13}.
If one were to substitute
\bd
\nmm{z}_1 - \nmm{x}_1 \leq 0
\ed
into the bound given in that theorem, the result would be
\bd
\nmp{ \xh - x } \leq \frac{1}{ k^{1 - 1/p} } \cdot \frac{2}{1 - \r}
[ (1+ \r )^2 \s_k(x,\nmm{\cdot}_1) + (3 + \r) \t \e ] .
\ed
The bound in \eqref{eq:2226} is better in that
$(1+\r)^2$ is replaced by $1 + 2 \r$, and $3+\r$ is
replaced by $3$.

\section{Numerical Example}\label{sec:exam}

In this section we illustrate the application of the bounds in 
\eqref{eq:41} and \eqref{eq:43}.
Specifically, we compare the number of measurements for group sparse
recovery as given in these bounds with the number for conventional
sparse recovery given in \eqref{eq:28g}.
As shown in \cite{Mahsa-TSP19}, unless $n$ is larger than about $10^5$,
the bound $m_C$ in \eqref{eq:28g} often exceeds $n$, which makes
``compressed'' sensing meaningless.
We study four different cases to illustrate the fact that
even with small groups of equal size, robust group sparse recovery can
require fewer samples than conventional sparse recovery.
Moreover, as the minimum size of the groups increases, the advantage is
even more on the side of group sparse recovery.
The reason for this phenomenon is that, as the minimum group size increases,
the total number of groups decreases.
Consequently, the cardinality of the number of group sparse sets decreases
fairly rapidly.
This is the quantity referred to as
$C(g,\phi)$
in the proof of Theorem \ref{thm:28}.

Specifically, we choose $n = 10^6$ and $k = 60$.
We use a sub-Gaussian random variable that satisfies the same rate
of decay as a standard normal variable, namely $\cb = 1/2$ in \eqref{eq:28b};
see \cite[Lemma 7.6]{FR13}.
To compute the RIP constant $\d$, we choose $t = 1.5$, which gives
$1/\sqrt{3} \approx 0.577$ as the upper limit for conventional sparse
recovery, as given in Theorem \ref{thm:CZ}.
We choose $\d_C = 0.5$, or about 85\% of the limit, as the RIP constant
for conventional sparsity.
In the case of GRIP, we compute the limit $\d_G$ as 85\% of the limit
$\bar{\d}_G$ given by \eqref{eq:315}.
Note that for different choices of group sizes, this threshold would
also be different.
Finally, for the failure probability $\xi$ we choose $10^{-9}$
for both conventional and group sparse recovery.
Table \ref{table:1} shows the number of samples needed by conventional sparsity
and group sparsity for various values of $\dmax, \dmin, g$.
The choice of the GRIP constant $\d_G$ for each choice of $\dmax, \dmin, g$
is also shown in the table.
Note that, from Theorem \ref{thm:1b}, when all groups have the same
size $d$, and both $n$ and $k$ are multiples of $d$, then the GRIP bound
$\d_G$ and RIP bound $\d_C$ are the same.

It is noteworthy that, even when the GRIP constant $\d_G$ that the matrix
$A$ is required to satisfy is smaller than the RIP constant $\d_C$,
the number of samples can be smaller in the case of group sparsity,
as happens in row 2 of the table.
This is because the number of group sparse sets is substantially smaller than
the number of sparse sets.
As a final example, we increased $k$ to $300$, and chose the group sizes
to be uniform at $50$ (thus leading to $20,000$ groups).
With this choice, $m_C = 1,464,244$, that is, \textit{more than} the
size of the vector, whereas $m_G = 393,153$.
Thus group sparse recovery is feasible when conventional sparse
recovery is not feasible.

\begin{table}
\bc
\btab{|c|c|c|c|c|c|c|}
\hline
$\dmax$ & $\dmin$ & $g$ & $\d_C$ & $\d_G$ & $m_C$ & $m_G$ \\
\hline
4 & 2 & 300,000 & 0.5 & 0.3750 & 335,862 & 545,935 \\
10 & 6 & 100,000 & 0.5 & 0.4091 & 335,862 & 296,499 \\
4 & 4 & 250,000 & 0.5 & 0.5000 & 335,862 & 162,492 \\
10 & 10 & 100,000 & 0.5 & 0.5000 & 335,862 & 124,505 \\
\hline
\etab
\ec
\caption{Comparison of number of measurements required in
conventional and group sparsity for various group sizes,
with $n = 10^6$ and $k = 60$.}
\label{table:1}
\end{table}

\section{Proofs of Main Results}\label{sec:proofs}

\subsection{Polytope Decomposition Lemma}\label{ssec:81}

The key to the results in \cite{CZ14} is Lemma 1.1 of that paper,
which the authors call the ``polytope decomposition lemma.''
In this subsection we generalize this lemma to the case of group sparsity.
Before presenting the lemma, we introduce a couple of terms.
Given a vector $v \in \R^n$, we define the \textbf{group support set of $v$},
denoted by $\Gsupp(v)$, as
\be\label{eq:31}
\Gsupp(v) := \{ j \in [g] : v_{G_j} \neq 0 \} .
\ee
Thus $\Gsupp(v)$ denotes the subset of the groups on which $v$ has
a nonzero support. 
Obviously $| \Gsupp(v)|$ is the number of distinct groups on which $v$
is supported.

\begin{lemma}\label{lemma:31}
Given a vector $v\in R^n$ such that,
\be\label{eq:33}
\nmm{ v_{G_j} }_1 \leq \al , \fa j \in [g],
\mbox{ and } \nmm{v}_1 \leq s \al 
\ee
for some integer $s$, there exist an integer $N$ and vectors 
$u_i, i \in [N]$ such that
\bit
\item $\supp(u_i) \seq \supp(v), \fa i \in [N]$.
\item $\nmm{u_i}_1 = \nmm{v}_1, \fa i \in [N]$.
\item $u_i$ is group $s \dmax$-sparse for each $i$, and finally
\item $v$ is a convex combination of $u_i , i \in [N]$.
\eit
\end{lemma}


\textbf{Remarks:} In the case of conventional sparsity,
each group $G_j$ consists of the singleton $\{ j \}$.
In this case the condition $\nmm{ v_{G_j} }_1 \leq \al , \fa j \in [g]$
reduces to $| v_j | \leq \al \fa j \in [n]$, or equivalently,
$\nmm{v}_\infty \leq \al$.
Moreover, $\dmax = 1$, in which case all vectors $u_i$ are $s$-sparse.
This is precisely \cite[Lemma 1.1]{CZ14}.

\textbf{Proof:}
The proof is by induction.
Define a subset of $\R^n$ as follows:
\bd
X := \{ v \in \R^n : \nmm{ v_{G_j} }_1 \leq \al \fa j \in [g] ,
\nmm{v}_1 \leq s \al \} .
\ed
To begin the inductive process, suppose $| \Gsupp(v) | \leq s$.
Then $v$ is itself $s \dmax$-sparse.
So we can take $N = 1$ and $u_1 = v$.
Now suppose that the lemma is true for all $v \in X$ such that
$| \Gsupp(v)| = r - 1$ where $r-1 \geq s$.
It is shown that the lemma is also true for all $v \in X$ satisfying
$| \Gsupp(v) | = r$.

Let $Q \seq [g]$ denote the index set $\{ j \in [g] : v_{G_j} \neq 0 \}$,
and observe that $|Q| = | \Gsupp(v) | = r$ by assumption.
Then $v$ can be expressed as $v = \sum_{j \in Q} v_{G_j}$.
Now arrange the vectors $v_{G_j}$ in decreasing order of their $\ell_1$-norm.
Denote the permuted vectors as $p_1$ through $p_r$.
Define $a_i := \nmm{p_i}_1$, and $\ph_i = (1/a_i) p_i$.
Then each $\ph_i$ has unit $\ell_1$-norm.
Moreover $a_i \geq a_{i+1}$ for all $i$, and
$v = \sum_{i=1}^r p_i = \sum_{i=1}^r a_i \ph_i$.
Also, because the $\ell_1$-norm is decomposable and the $p_i$ have
nonoverlapping support sets, it follows that
$\nmm{v}_1 = \sum_{i=1}^r a_i$.

Now define a set
\bd
D := \{ \beta \in [r-1] : \sum_{i=\beta}^r a_\beta \leq (r - \beta) \al \} .
\ed
Then $1 \in D$ because
\bd
\sum_{i=1}^r a_i = \nmm{v}_1 \leq s \al \leq (r-1) \al .
\ed
Therefore $D$ is nonempty.
Now, by a slight abuse of notation, let $\beta$ again denote the
largest element of the set $D$.
This implies that
\be\label{eq:34}
\sum_{i=\beta}^r a_i \leq (r - \beta) \al ,
\sum_{i=\beta+1}^r a_i > (r - \beta - 1) \al .
\ee
Define the constants
\bd
b_t := \frac{1}{r - \beta} \sum_{i=\beta}^r a_i - a_t , \beta \leq t \leq r .
\ed
Since the first term on the right side is independent of $t$, and
$a_{t+1} \leq a_t$, it follows that $b_{t+1} \geq b_t$.
Also
\begin{eqnarray*}
b_\beta & = & \frac{1}{r-\beta} \sum_{i=\beta}^r a_i - a_\beta \\
& = & \frac{1}{r-\beta} \sum_{i=\beta+1}^r a_i - \frac{r - \beta - 1}{r - \beta}
a_\beta \\
& \geq & \frac{1}{r-\beta} \left[ \sum_{i=\beta+1}^r a_i - (r - \beta - 1) \al 
\right ] > 0 ,
\end{eqnarray*}
where the last two steps follow from $a_i \leq \al$ for all $i$, and
from the second inequality in \eqref{eq:34}.
Also, it is easy to verify that
\be\label{eq:36}
\sum_{i=\beta}^{r} a_i = (r-\beta) \sum_{i=\beta}^{r}b_i
\ee
Next, for $t = \beta , \ldots , r$, define
\be\label{eq:37}
w_t := \sum_{i=1}^{\beta-1} a_i \ph_i + \left( \sum_{i=\beta}^r b_i \right)
\sum_{i=\beta, i \neq t}^r \ph_i ,
\l_t := \frac{b_t}{\sum_{i=\beta}^r b_i } .
\ee
Now observe that 
\bd
0 < \l_t < 1, \sum_{t=\beta}^r \l_t = 1 ,
\mbox{ and } v = \sum_{t=\beta}^r \l_t w_t .
\ed
Next, $\supp(w_t) \seq \supp(v)$ for all $t$.
Moreover, $| \Gsupp(w_t) | \leq r-1$ for all $t$, because the corresponding
term $\ph_t$ is missing from the summation in \eqref{eq:37}.
Also, note that each $\ph_i$ has unit $\ell_1$-norm.
Therefore, for each $t$ between $\beta$ and $r$, we have that
\begin{eqnarray*}
\nmm{w_t}_1 & = & \sum_{i=1}^{\beta-1} a_i + (r - \beta) \sum_{i=\beta}^r b_i \\
& = & \sum_{i=1}^{\beta-1} a_i + \sum_{i=\beta}^{r} a_i 
= \sum_{i=1}^r a_i = \nmm{v}_1 .
\end{eqnarray*}
Therefore each $w_t \in X$.
By the inductive assumption, each $w_t$ has a convex decomposition as
in the statement of the lemma.
It follows that $v$ is also a convex combination as in the statement of
the lemma.
This completes the inductive step.
\halmos

\begin{lemma}\label{lemma:32}
Let $u_i , i \in [N]$ be the vectors in the convex combination of
Lemma \ref{lemma:31}.
Then
\be\label{eq:39}
\nmeusq{u_i} \leq \frac{s \dmax}{\dmin} \al^2 , \fa i \in [N] .
\ee
\end{lemma}

\textbf{Proof:}
Fix the index $i \in [N]$.
Define the index set
\bd
B_i := \{ j \in [g] : (u_i)_{G_j} \neq 0 \} .
\ed
Let $c_i = |B_i|$.
Because $u_i$ is $s \dmax$-sparse, it follows that
$c_i \leq \frac{s \dmax}{\dmin}$.
Moreover, for each index $j \in B_i$, we have that
\bd
\nmeu{ (u_i)_{G_j} } \leq \nmm{ (u_i)_{G_j} }_1 \leq \nmm{u_i}_1 = \al .
\ed
Now observe that
\bd
u_i = \sum_{j \in B_i} (u_i)_{G_j} .
\ed
Next, note that the various vectors $(u_i)_{G_j}$ are supported on
disjoint sets.
Therefore
\bd
\nmeusq{u_i} = \sum_{j \in B_i} \nmeusq{(u_i)_{G_j}} .
\ed
Since there are $c_i$ terms in the above summation,
and each term is no larger than $\al^2$,
it follows that
\bd
\nmeusq{u_i} \leq c_i \al^2 \leq \frac{s \dmax}{\dmin} \al^2 ,
\ed
which is the desired conclusion \eqref{eq:39}.
\halmos

\subsection{Group Robust Null Space Property}\label{ssec:82}


\textbf{Proof of Theorem \ref{thm:1}:}
Recall the constants $\nu, a, b, c_G, \r_G, \t_G$ defined in the statement of
Theorem \ref{thm:1}.
We will make use of these constants in the proof.
Let $h \in \R^n$ be arbitrary.
The objective is to establish that the inequality \eqref{eq:314}
is satisfied with $\r_G,\t_G$ defined as above.

Let $h_{\L_{0}}, h_{\L_{1}}, h_{\L_{2}}, \ldots ,
h_{\L_{s}}$ be an optimal group-$k$-sparse decomposition of $h$.
This means the following:
First,
\bd
h_{\L_0} = \argmin_{ \supp(z) \in \GkS} \nmm{ x - z} .
\ed
Next, for $i \geq 1$,
\bd
h_{\L_i} = \argmin_{ \supp(z) \in \GkS} \left\nm x - \sum_{j=0}^{i-1}
h_{\L_j} - z \right\nm .
\ed
Now denote $h_{\L_0^c} =h^*$.
Define sets $S_1$ and $S_2$ as follows:
\bd
S_1 = \left\{ j : \nmm{ h^*_{G_j} }_1 > 
\frac{ \nmm{ \hloc }_1}{k(t-1)}, \fa j \in [g] \right\} ,
\ed
\bd
S_2 = \left\{ j : \nmm {h^*_{G_j} }_1 \leq 
\frac{ \nmm{ \hloc }_1}{k(t-1)}, \fa j \in [g] \right\} .
\ed
Let $GS_1 = \cup_{j \in S_1} G_j$ and $GS_2 = \cup_{j \in S_2} G_j$.
Now define
\bd
h^{(0)} = \hlo , h^{(1)} = h^*_{GS_1}, h^{(2)} = h^*_{GS_2} .
\ed
Then we have
\bd
\hloc = h^* =  h^*_{GS_1} + h^*_{GS_2} =  h^{(1)}+h^{(2)} .
\ed
Let  $r = |S_1|$, and note that $r \leq k(t-1)$.
This is because, by the manner in which we defined the set $S_1$,
it follows that 
\bd
\nmm{ \hloc }_1 \geq \nmm{ h^{(1)} }_1 >
r \frac{ \nmm{ \hloc }_1}{k(t-1)} .
\ed
Next we establish upper bound on $\nmm{ h^{(2)} }_1$.
Because of the definition of set $S_1$, it follows that
\be\label{eq:316}
\nmm{ h^{(1)} }_1 \geq r \frac{ \nmm{ \hloc }_1}{k(t-1)} .
\ee
Therefore
\beq
\nmm{ h^{(2)} }_1 & = &  \nmm{ \hloc }_1 - \nmm{ h^{(1)} }_1  \nonumber \\
& \leq & \nmm{ \hloc }_1 - r 
\frac{ \nmm{ \hloc }_1 }{k(t-1)} \nonumber \\
& = & [k(t-1)- r ] \frac{ \nmm{ \hloc }_1 }{k(t-1)} . \label{eq:317}
\eeq
By the definition of set $S_2$
\be\label{eq:318}
\nmm{ h^{(2)}_{{G}_j} }_1 \leq 
\frac{ \nmm{ \hloc }_1 }{k(t-1)} , \fa j \in [g] .
\ee
From \eqref{eq:317} and \eqref{eq:318}, we see that the vector
$h^{(2)}$ satisfies the hypotheses of Lemma \ref{lemma:31} with
\bd
\al = \frac{ \nmm{ \hloc }_1 }{k(t-1)},
s =  k(t-1)- r  .
\ed
Therefore we can apply Lemma \ref{lemma:31} to $h^{(2)}$.
So $h^{(2)}$ can be represented as 
\be\label{eq:318a}
h^{(2)} = \sum_{i=1}^N \l_i u_i ,
\ee
where each $u_i$ is group $(k(t-1)- r) \dmax$-sparse,
$h^{(1)}$ is group $(r \dmax)$-sparse, and $h^{(0)}$ is group $k$-sparse.
Therefore $u_i+ h^{(1)}+\ h^{(0)}$ has group sparsity no larger than
\begin{eqnarray*}
k + r \dmax + (k(t-1) - r) \dmax & = & k [ 1 + (t-1) \dmax ] \\
& = & \kb 
\end{eqnarray*}
for each $i \in [N]$.
Now let, for all $i \in [N]$,
\bd
x_i = \frac{1}{2}\Big(h^{(0)}+h^{(1)}\Big)+\ \frac{\nu}{2}u_i ,
\ed
\bd
z_i =  \frac{1-2\nu}{2}\Big(h^{(0)}+h^{(1)}\Big)-\ \frac{\nu}{2}u_i ,
\ed
\bd
\g =  x_i + z_i =  (1-\nu) \left( h^{(0)}+h^{(1)} \right) ,
\ed
\bd
\beta_i =  x_i - z_i =  \nu \left (h^{(0)}+h^{(1)}+\ u_i \right) .
\ed
Then
\beq
\sum_{i=1}^{N}\l_i \IP { A\g } { A\beta_i } 
& = &  \left\langle A\g , A \sum_{i=1}^{N} \l_i \beta_i \right\rangle 
\nonumber \\
& = & \nu(1-\nu) \IP { A(h^{(0)}+h^{(1)}) } { Ah } , \label{eq:319}
\eeq
where we make use of \eqref{eq:318a} and the fact that 
$h^{(0)}+h^{(1)}+h^{(2)} = h$.
However, for each index set $i$, we have that
\begin{eqnarray*}
\IP { A \g }{ A \beta_i } & = &  \IP { Ax_i+Az_i }{ Ax_i-Az_i } \\
& = &  \nmeusq{Ax_i} - \nmeusq{Az_i} .
\end{eqnarray*}
Therefore it follows that
\bd
\sum_{i=1}^{N} \l_i (\nmeusq{Ax_i} - \nmeusq{Az_i})
=  \nu (1-\nu) \left\langle A(h^{(0)}+h^{(1)}) , Ah \right\rangle ,
\ed
\begin{eqnarray*}
\sum_{i=1}^{N} \l_i \nmeusq{Ax_i} 
& = & \sum_{i=1}^{N}\l_i \nmeusq{Az_i} \\
& + &\nu(1-\nu)\Big\langle A(h^{(0)}+h^{(1)})\ ,\ Ah\Big\rangle .
\end{eqnarray*}
Since $x_i,\ z_i,\ (h^{(0)}+h^{(1)})$ are all group $\kb$-sparse,
it follows from the GRIP and Schwarz' inequality that
\begin{eqnarray*}
(1-\d) \sum_{i=1}^{N} \l_i \nmeusq{ x_i } & \leq & 
(1+\d) \sum_{i=1}^{N} \l_i \nmeusq{ z_i } \nonumber \\
& + & \nu(1-\nu) \nmeu{ A(h^{(0)}+h^{(1)}) } \cdot \nmeu{ Ah } .
\end{eqnarray*}
Since $h^{(0)},\ h^{(1)}$ and $u_i$ have disjoint support sets,
it follows that, for all $i \in [N]$, we have
\bd
\nmeusq{x_i} =  0.25\left( \nmeusq{(h^{(0)} + h^{(1)})} + \nu^2
\nmeusq{u_i} \right),
\ed
\bd
\nmeusq{z_i} =  0.25 \left[ (1-2\nu)^2 \nmeusq{(h^{(0)}+h^{(1)})}
+ \nu^2 \nmeusq{u_i} \right],
\ed
Substituting these relationships,
multiplying both sides by 4, and noting that $\sum_{i=1}^{N} \l_i =  1$,
leads to
\begin{eqnarray*}
(1-\d) & \cdot & \left[ \nmeusq{ (h^{(0)}+h^{(1)}) }
+ \nu^2 \sum_{i=1}^{N} \l_i \nmeusq{ u_i } \right] \\
& \leq & (1+\d) \left[ (1-2\nu)^2 \nmeusq{ (h^{(0)}+h^{(1)}) } \right. \\
& + & \left. \nu^2 \sum_{i=1}^{N} \l_i \nmeusq{ u_i } \right] \\
& +  & 4 \nu (1-\nu) \nmeu{ A(h^{(0)}+h^{(1)}) } \cdot \nmeu{ Ah } ,
\end{eqnarray*}
or upon rearranging,
\begin{eqnarray*}
\nmeusq{ (h^{(0)}+h^{(1)}) } & \cdot &
[(1-\d) - (1+\d) (1-2\nu)^2]  \\
& \leq & 2 \d \nu^2 \sum_{i=1}^{N} \l_i \nmeusq{ u_i }  \\
& + & 4 \nu (1-\nu) \nmeu{ A(h^{(0)}+h^{(1)}) } \cdot \nmeu{ Ah } .
\end{eqnarray*}
Recall that  
\bd
\al =  \frac{\nmm{ \hloc }_1}{k(t-1)} ,
s =  k(t-1)- r .
\ed
Substituting these values into \eqref{eq:39}, we get that 
\begin{eqnarray*}
\nmeusq{ u_i } & \leq & [ k(t-1)- r ] \frac{ \dmax }{ \dmin }
\frac{\nmm{ \hloc }_1^2}{k^2(t-1)^2} \\
& \leq & k(t-1) \frac{ \dmax }{ \dmin }
\frac{\nmm{ \hloc }_1^2}{k^2(t-1)^2} \\
& = &  \frac{\dmax}{\dmin}\ \frac{\nmm{ \hloc }_1^2}{k(t-1)} .
\end{eqnarray*}
Substituting this bound, which is independent of $i$, into
the above inequality, and noting that $\sum_{i = 1}^N \l_i = 1$, we get
\begin{eqnarray*}
\nmeusq{ (h^{(0)} & + &h^{(1)}) } [(1-\d)-(1+\d)(1-2\nu)^2] \\
& \leq & \frac{2\d \nu^2 \dmax}{\dmin} \frac{\nmm{ \hloc }_1^2}{k(t-1)}\\
& + & 4\nu(1-\nu) \sqrt{1+\d} \nmeu{ (h^{(0)}+h^{(1)}) } \cdot \nmeu{ Ah } .
\end{eqnarray*}
Denote $\nmeu{ (h^{(0)}+h^{(1)}) }$ by $f$ and invoke the definition of 
the constants $a,b,c$ from \eqref{eq:311} and \eqref{eq:312}.
This gives

\bd
4 f^2 a^2 \leq 4c^2 \frac{ \nmm{ \hloc }_1^2}{k} + 4bf \nmeu{Ah} ,
\ed
or after dividing both the sides by 4 and rearranging,
\bd
f^2 a^2 - bf \nmeu{ Ah } \leq c^2 \frac{\nmm{ \hloc }_1^2}{k} .
\ed
The next step is to complete the square on left side of the above inequality.
This gives
\bd
f^2 a^2 - bf \nmeu{ Ah} + \frac{b^2}{4a^2} \nmeusq{ Ah } \leq
\frac{b^2}{4a^2}\nmeusq{ Ah } + c^2 \frac{\nmm{ \hloc }_1^2}{k} ,
\ed
or equivalently,
\bd
\left[ af - \frac{b}{2a} \nmeu{ Ah }\right]^2
\leq \frac{b^2}{4a^2} \nmeusq{ Ah } + c^2 \frac{\nmm{ \hloc }_1^2}{k} .
\ed
Taking the square root on both sides, and using the obvious inequality that
$\sqrt{x^2+y^2} \leq x+y$ whenever $x,y\geq 0$, leads to 
\bd
af-(b/2a) \nmeu{ Ah } \leq (b/2a) \nmeu{ Ah }
+ c \frac{ \nmm{ \hloc }_1}{\sqrt{k}} ,
\ed
or upon rearranging and replacing $f$ by $\nmeu{ (h^{(0)}+h^{(1)})}$,
\bd
a \nmeu{ (h^{(0)}+h^{(1)}) } \leq (b/a) \nmeu{ Ah }
+ c \frac{ \nmm{ \hloc }_1}{\sqrt{k}} .
\ed
Dividing both the sides by $a$ and observing that $h_{\L_0} =  h^{(0)}$ and 
\bd
\nmeu{ h^{(0)} } \leq \nmeu{ (h^{(0)}+h^{(1)}) } ,
\ed
we get
\begin{eqnarray*}
\nmeu{ \hlo } & \leq & \nmeu{ (h^{(0)}+h^{(1)}) } \leq
\frac{b}{a^2} \nmeu{ Ah } + \frac{c}{a} \frac{\nmm{ \hloc }_1}{\sqrt{k}} \\
& = & \frac{b\sqrt{k}}{a^2\sqrt{k}} \nmeu{Ah} +
\frac{c}{a} \frac{\nmm{ \hloc}_1}{\sqrt{k}} .
\end{eqnarray*}
This inequality is of the form \eqref{eq:314} with $\r_G, \t_G$ given as in
\eqref{eq:313}.

The proof is therefore complete once it is shown that $\r_G = c_G/a < 1$
if and only if $\d_{G,\kb} < \bar{\d}_G$.
Towards this end, define
\bd
\al = \frac{\nu^2 \dmax}{2(t-1) \dmin} .
\ed
Then $c_G^2 = \al \d$.
Next, observe that $c_G < a$ if and only if $c_G^2 < a^2.$
Now we can invoke the definitions of $a$ and $c_G$ from \eqref{eq:311}
and \eqref{eq:312a}, which leads to
\begin{eqnarray*}
c_G^2 < a^2 & \iff & \al \d < \nu(1-\nu)-\delta(0.5-\nu(1 - \nu)) \\
& \iff & \d [ \al + 0.5-\nu(1 - \nu) ] < \nu(1-\nu) \\
& \iff & \d < \bar{\d}_G ,
\end{eqnarray*}
where $\d$ is shorthand for $\d_{G,\kb}$.
\halmos

\textbf{Proof of Theorem \ref{thm:1b}:}
If all groups have the same size, then $\dmax = \dmin = d$, and the bound
\eqref{eq:315} on the restricted isometry constant $\d_{G,\kb}$ becomes
\be\label{eq:313bb}
\bar{\d} = \nu (1 - \nu) \left( \frac{ \nu^2 }{2(t-1) } +
0.5 - \nu + \nu^2 \right)^{-1} .
\ee
The objective is to show that $\bar{\d} = \sqrt{(t-1)/t} =: \psi$ say.
From \eqref{eq:313bb}, the statement that $\bar{\d} = \psi$ is equivalent to
\bd
\nu (1 - \nu) = \frac{\psi}{2} \left( \frac{ \nu^2 }{(t-1) } + 1 \right)
- \psi \nu (1 - \nu) ,
\ed
which in turn is equivalent to
\be\label{eq:313b}
2 ( 1 + \psi ) \nu (1 - \nu) = \psi \left( \frac{ \nu^2 }{(t-1) } + 1 \right) .
\ee
Now note that, from the definition \eqref{eq:310} of the constant $\nu$,
it follows that
\bd
\nu = t \sqrt{ \frac{t-1}{t} } - (t-1) = t \psi -t +1 = 1 - t(1 - \psi) ,
\ed
and
\bd
1 - \nu = t(1 - \psi) .
\ed
Therefore the left side of \eqref{eq:313b} becomes
\bd
2 ( 1 + \psi ) \nu (1 - \nu) = 2t (1 - \psi^2) \nu .
\ed
However
\bd
t (1 - \psi^2) = t \left( 1 - \frac{t-1}{t} \right) = 1 .
\ed
Therefore the left side of \eqref{eq:313b} equals $2 \nu$.
The proof is therefore complete if it can be shown that the right side of
\eqref{eq:313b} also equals $2 \nu$.
Towards this end, note that
\bd
\nu^2 = t(t-1) -2(t-1)\sqrt{t(t-1)} + (t-1)^2 ,
\ed
\bd
\frac{ \nu^2 }{t-1} = t - 2 \sqrt{t(t-1)} + t-1  ,
\ed
\bd
\frac{ \nu^2 }{t-1} + 1 = 2 (t - \sqrt{t(t-1)} ) ,
\ed
and finally
\begin{eqnarray*}
\psi \left( \frac{ \nu^2 }{t-1} + 1 \right) 
	& = & 2 \sqrt{ \frac{t-1}{t} } (t - \sqrt{t(t-1)} ) \\
	& = & 2 [ \sqrt{t(t-1)} - (t-1) ] = 2 \nu .
\end{eqnarray*}
\halmos

\textbf{Proof of Theorem \ref{thm:1a}:}
This consists of the observation that if $\dmax = \dmin = 1$, then
$\kb = [1 + (t-1) \dmax] k = tk$.
\halmos

\subsection{Error Bounds on the Recovered Vector}\label{ssec:83}

\textbf{Proof of Theorem \ref{thm:2}:}
Define $\xh$ as in \eqref{eq:11a}, and let $h = \xh - x$ denote the
residual error.
Then by definition we have that $\nmm{\xh}_1 \leq \nmm{x}_1$.
Let $x_{S_0},x_{S_1}, \ldots , x_{S_b}$ be an
optimal  group $k$-sparse decompostion of $x$. 
Then
\bd
\nmm{x_{S_0^c}+h_{S_0^c}}_1+\nmm{x_{S_0}+h_{S_0}}_1 \leq \nmm{x_{S_0^c}}_1+\nmm{x_{S_0}}_1 .
\ed
Applying triangle inequality twice to the left hand side of the above
inequality, we get
\bd
\nmm{x_{S_0}}_1-\nmm{h_{S_0}}_1-\nmm{x_{S_0^c}}_1+\nmm{h_{S_0^c}}_1\leq\nmm{x_{S_0^c}}_1+\nmm{x_{S_0}}_1 .
\ed
Cancelling the common term $\nmm{x_{S_0}}_1$ and denoting $\nmm{x_{S_0^c}}$ by
$\sigma_{k,{G}}(x,\nmm{\cdot}_1) = \sigma_{k,{G}}$, we get
\be\label{eq:341}
\nmm{h_{S_0^c}}_1-\nmm{h_{S_0}}_1\leq 2\sigma_{k,{G}}
\ee
Now let
$h_{\L_{0}},h_{\L_{1}}, \ldots , h_{\L_{s}}$ be an optimal group $k$-sparse
decomposition of $h$. 
Then 
\bd
\nmm{h_{\L_{0}}}_1\geq\ \nmm{h_{S_0}}_1, \mbox{ and } \nmm{h_{\L_{0}^c}}_1 \leq  \nmm{h_{S_0^c}}_1 .
\ed
Using the above facts in \eqref{eq:341}, we get 
\begin{equation}\label{eq:342}
\nmm{h_{\L_{0}^c}}_1 - \nmm{h_{\L_{0}}}_1 \leq 2\sigma_{k,{G}} .
\end{equation}
Next, because both $x$ and $\xh$ are feasible for the optimization
problem in \eqref{eq:11a}, we get
\bd
\nmeu{Ah} =  \nmeu{(A\hat{x}-y)-(A{x}-y)} \leq  2\epsilon .
\ed
Using the inequality \eqref{eq:314a} and the above fact, we have that
\be\label{eq:343}
\nmm{h_{\L_{0}}}_1 \leq  {\r_G}\nmm{h_{\L_{0}^c}}_1\ +\ 2{\t_G}\epsilon.
\ee
Now the two inequalities \eqref{eq:342} and \eqref{eq:343} can
be neatly expressed in the form
\be\label{eq:344}
\begin{bmatrix}
1 &  -1\\
-\r_G & 1
\end{bmatrix}
\begin{bmatrix}
\nmm{h_{\L_{0}^c}}_1\\
\nmm{h_{\L_{0}}}_1
\end{bmatrix}
\leq
\begin{bmatrix}
2\sigma_{k,{G}}\\
2\t_G\epsilon
\end{bmatrix} .
\ee
Let the $M$ denote the coefficient matrix on the left hand side.
Then, because $\r_G < 1$, it follows that all elements of
\bd
M^{-1} =
\frac{1}{1-\r_G}\begin{bmatrix}
1 &  1\\
\r_G & 1
\end{bmatrix}
\ed
are positive.
Therefore we can multiply both the sides of \eqref{eq:344} by $M^{-1}$,
which gives
\begin{align}
\begin{bmatrix}
\nmm{h_{\L_{0}^c}}_1\\
\nmm{h_{\L_{0}}}_1
\end{bmatrix} \leq  & \frac{1}{1-\r_G}\begin{bmatrix}
1 &  1\\
\r_G & 1
\end{bmatrix}\begin{bmatrix}
2\sigma_{k,{G}}\\
2\t_G\epsilon
\end{bmatrix}\nonumber \\ 
= &\frac{2}{1-\r_G}\begin{bmatrix}
(\sigma_{k,{G}}+\t_G\epsilon)\\
(\r_G\sigma_{k,{G}}+\t_G\epsilon)
\end{bmatrix}
\label{eq:345}
\end{align}
Finally using the triangle inequality, we get
\begin{align}
\nmm{h}_1&\leq\nmm{h_{\L_{0}^c}}_1+\ \nmm{h_{\L_{0}}}_1\nonumber\\
& = \big[1\ \ \ \ 1\big]\begin{bmatrix}
\nmm{h_{\L_{0}^c}}_1\\
\nmm{h_{\L_{0}}}_1
\end{bmatrix}\nonumber\\
& \leq  \frac{2}{1-\r_G}[(1+\r_G)\sigma_{k,{G}}+\ 2\t_G\epsilon]
\nonumber
\end{align}
This is the same as \eqref{eq:350}.

Next we derive bounds on $\nmp{h}$ for $p \in [1,2]$.
From the triangle inequality,
\begin{equation}\label{eq:346}
\nmp{h} \leq  \nmp{h_{\L_{0}}}+\ \nmp{h_{\L_{0}^c}} .
\end{equation}
Now we will obtain the upper bound for both of the terms in right hand side of
\eqref{eq:346}.
It is easy to show that
\be\label{eq:345a}
\nmp{h_{\L_{0}^c}} \leq  \nmm{h_{\L_{0}^c}}_1 .
\ee
Next, it is a ready consequence of  H\"{o}lder's inequality that 
\bd
\nmp{h_{\L_{0}}} \leq  k^{1/p-1/2}\ \nmeu{h_{\L_{0}}} .
\ed
Using the above fact and the $\ell_2$-GRNS property \eqref{eq:314},
together with $\nmeu{Ah} \leq 2\e$, we get 
\be\label{eq:348}
\nmeu{\hlo} \leq \frac{\r_G}{\sqrt{k}} \nmm{\hloc}_1 + \frac{2 \t_G \e}{\sqrt{k}} .
\ee
Therefore
\be\label{eq:348a}
\nmp{\hlo} \leq \frac{1}{k^{1-1/p}}
[ \r_G \nmm{\hloc}_1 + 2 \t_G \e ] .
\ee
Combining \eqref{eq:345a} and \eqref{eq:348a} leads to
\be\label{eq:348b}
\nmp{h} \leq \left( 1 + \frac{\r_G}{k^{1-1/p}} \right) \nmm{\hloc}_1 
+ \frac{2 \t_G \e}{k^{1-1/p}} .
\ee
Now we can substitute the upper bound for $\nmm{\hloc}_1$ obtained from
\eqref{eq:345}, namely
\bd
\nmm{\hloc}_1 \leq \frac{2}{1-\r_G} ( \s_{k,G} + \t_G \e ) .
\ed
Substituting this bound into \eqref{eq:348b} leads finally to the bound
\beq
\nmp{h} & \leq & \frac{2}{1-\r_G} \left( 1 + \frac{\r_G}{k^{1-1/p}}
\right) \s_{k,G}
\nonumber \\
& + & \left[ \frac{2}{1-\r_G} \left( 1 + \frac{\r_G}{k^{1-1/p}} \right)
+ \frac{2}{k^{1-1/p}} \right] \t_G \e . \nonumber
\eeq
This is precisely \eqref{eq:2226}.
\halmos

\textbf{Proof of Theorem \ref{thm:26}:}
In this case the optimal group $k$-sparse decomposition becomes
just the conventional optimal $k$-sparse decomposition.
To prove \eqref{eq:2226}, we proceed as above.
However, instead of \eqref{eq:345a},
we use the inequality from \cite[Theorem 2.5]{FR13}, namely
\be\label{eq:348c}
\nmp{\hloc} = \s_k(h,\nmp{\cdot}) \leq \frac{1}{k^{1-1/p}} \nmm{h}_1 .
\ee
Note that an analogous inequality does not exist for \textit{group}
$k$-sparse decompositions.
Now we merely substitute the bound from \eqref{eq:348c} instead of the
bound from \eqref{eq:345} into \eqref{eq:348b}; this leads to \eqref{eq:2226}.
\halmos

\subsection{Sample Complexity Estimates}\label{ssec:84}


\textbf{Proof of Theorem \ref{thm:28}:}
The proof is a fairly straight-forward adaptation of that of
\cite[Theorem 9.9, p.\ 276]{FR13}, and \cite[Theorem 9.11, p.\ 278]{FR13}.
By assumption, \eqref{eq:28c} holds for \textit{every fixed} $u \in \R^n$.
By applying compactness arguments, it is shown in the cited proofs that,
for \textit{every fixed} subset $S$ of cardinality $s$ in $[n]$, 
the corresponding $m \times s$ submatrix $A_S$ satisfies the bound
\be\label{eq:51}
\s_{{\rm min}}(A_S^T A_S) \leq 1 - \d , \mbox{ and }
\s_{{\rm max}}(A_S^T A_S) \geq 1 + \d , 
\ee
with probability $\geq 1 - \th$, where
\be\label{eq:52}
\th = 2 \left( 1 + \frac{2}{\r} \right)^s \exp( - \ct (1 - 2 \r)^2 \d^2 m ) .
\ee
The above bound holds for \textit{all} constants $\r$.
See \cite[(9.12)]{FR13}.
Now by enumerating \textit{all possible subsets} of $[n]$ of cardinality
$s$, we get that the quantity
\be\label{eq:53}
\xi = \left( \ba{c} n \\ s \ea \right) \th =: C(n,s) \th
\ee
is an upper bound on the probability that $A$ \textit{fails} to satisfy
the RIP of order $s$ with constant $\d$.
By choosing $\r = 2/(e^{7/2} - 1)$, we get the bound in the last (unnumbered)
equation in the proof of \cite[Theorem 9.11, p.\ 278]{FR13}.
Substituting $s = tk$ gives the bound in \eqref{eq:28g}.

In the case of group sparsity,
define $\phi$ as in \eqref{eq:41a}, and observe that
any group $\kb$-subset of $[n]$ can be the union of no more than $\phi$
sets from the collection $G_1 , \ldots , G_g$.
Therefore the number of group $\kb$-sparse subsets of $[n]$ is bounded by
the combinatorial parameter $C(g,\phi)$.
Therefore, replacing $C(n,tk)$ by $C(g,\phi)$,
or what is the same, changing $n$ to $g$ and $tk$ to $\phi$ in
\eqref{eq:28g}, gives the desired sample complexity estimate \eqref{eq:41}.
\halmos


\textbf{Proof of Theorem \ref{thm:29}:}
The bound in \eqref{eq:43} follows readily from that in \eqref{eq:41}
by substituting $n = gd$, $k = ld$, and $\phi = l[1+(t-1)d]$.
The following fact can be easily proved using undergraduate calculus:
The function $x \mapsto x \ln (eg/x)$ is strictly increasing for $x < g$.
With $n = gd, k = ld$ where $d > 1$, we get
\bd
\kb = [ 1 + (t-1)d ] k = [ 1 + (t-1)d ] ld ,
\ed
\bd
\phi = \frac{\kb}{d} = [ 1 + (t-1)d] l < [ d + (t-1)d ] l = tld .
\ed
Now compare the right sides of \eqref{eq:43} and \eqref{eq:42}.
First, because $\phi < tld$, we infer that
\bd
\phi \ln \frac{eg}{\phi} < tld \ln \frac{eg}{tld}
< tld \ln \frac{eg}{tl}
\ed
because $d > 1$.
So the first term in \eqref{eq:43} is smaller than the corresponding
term in \eqref{eq:42} if $tld < g$.
The second term in \eqref{eq:43} is smaller than the corresponding
term in \eqref{eq:42} because $\phi < tld$, and the third terms are
the same in both equations.
Therefore $m_G < m_C$ if $tld < g$.
\halmos

\section{Conclusions}\label{sec:conc}

In this paper we have shown that the $\ell_1$-norm is the convex relaxation
of two commonly used group sparsity indices.
Therefore $\ell_1$-norm minimization can be used
for recovering group sparse vectors, and not just for recovering
conventionally sparse vectors.
We have presented sufficient conditions for $\ell_1$-norm
minimization to achieve robust group sparse recovery, which are
less conservative than currently available results, based on minimizing
a group LASSO type of norm.
We achieved this by introducing a group version of the robust null space
property, and showing that GRNSP implies a group restricted isometry property.
This relationship is new even for conventional sparsity.
When specialized to conventional sparsity, our conditions for group sparse
recovery reduce to some known ``best possible'' bounds proved earlier.
We have also derived bounds for the $\ell_p$-norm of the residual error
between the true vector $x$ and its approximation $\xh$, for all $p \in [1,2]$.
These bounds are new even for conventional sparsity and of course also for
group sparsity.
For the case where the measurement matrix consists of random sub-Gaussian
samples, we have derived bounds for the number of samples that suffice
for group sparse recovery.
When all groups have the same size, our bounds are the same as known
bounds, while our bounds are less conservative when group sizes are not
all equal.
We have illustrated our approach through numerical examples.

There are two interesting avenues of research that are worth pursuing.
First, our results extend those in \cite{CZ14} to conventional
and group sparsity in terms of the RIP (or group RIP) coefficient
of order $\d_{tk}$ when $t > 1$.
It appears worthwhile to see whether the approach presented here can
also be applied to the case $t \in (0,4/3)$ studied in \cite{Zhang-Li-TIT18}.
Second, there is yet another model of sparsity referred to as ``joint''
sparsity in \cite{Elham-Vidal-CVPR11,Elham-Vidal-TSP12}.
It would be worthwhile to study whether 
the problem of recovering jointly sparse vector is amenable to the
approach presented here.

\section*{Acknowledgements}

We thank three anonymous reviewers for very helpful suggestions on a previous
version of the paper that have
greatly improved the contents and readability of the paper.

\bibliographystyle{IEEEtran}

\bibliography{Comp-Sens}

\end{document}